\def\year{2020}\relax
\documentclass[letterpaper]{article} 
\usepackage{aaai20}  
\usepackage{times}  
\usepackage{helvet} 
\usepackage{courier}  
\usepackage[hyphens]{url}  
\usepackage{graphicx} 
\usepackage{graphicx}  
\usepackage{booktabs}

\title{Towards Using Social HRI for Improving Children's Healthcare Experiences}

\author{Mary Ellen Foster \\
  School of Computing Science \\
  University of Glasgow \\
  Glasgow, Scotland, United Kingdom \\
  \texttt{\small MaryEllen.Foster@glasgow.ac.uk}
\And
  Ronald P. A. Petrick \\
  Edinburgh Centre for Robotics \\
  Heriot-Watt University \\
  Edinburgh, Scotland, United Kingdom \\
  \texttt{\small R.Petrick@hw.ac.uk} \\
}

\begin{document}

\maketitle

\begin{abstract}
This paper describes a new research project that aims to develop a 
social robot designed to help children cope with painful and distressing medical
procedures in a clinical setting. While robots have previously been trialled for this
task, with promising initial results, the systems have tended to be teleoperated,
limiting their flexibility and robustness. This project will use epistemic planning
techniques as a core component for action selection in the robot system, in order to
generate plans that include physical, sensory, and social actions for interacting
with humans.  The robot will operate in a task environment where appropriate and safe
interaction with children, parents/caregivers, and healthcare professionals is
required.  In addition to addressing the core technical challenge of building an autonomous 
social robot, the project will incorporate co-design techniques involving all
participant groups, and the final robot system will be evaluated in a two-site
clinical trial.
\end{abstract}

\section{Introduction}

Children regularly experience pain and distress in medical situations that can
produce both short-term (e.g., fear, distress, inability to perform procedures) and
long-term (e.g., needle phobia, anxiety) negative effects
\cite{Stevens.etal:2011}. While a range of techniques have been
demonstrated as effective methods for managing such situations (e.g., breathing
exercises, distraction techniques, cognitive-behavioural interactions
\cite{Chambers.etal:2009}), delivered through a variety of means (e.g.,
distraction cards, kaleidoscopes, music, and virtual reality games),
recent studies have also explored the use of social
robots as a tool for managing child pain and distress during medical procedures
\cite{Ali.etal:2019,Trost.etal:2019}.

One significant drawback in many of these past studies is that the robots tend to be
teleoperated or use purely scripted behaviours with limited autonomy and
responsiveness, thereby reducing the overall flexibility and robustness of the
system. Such approaches also limit the ability to personalise or adapt the system to
different user groups (e.g., children, parents of children undergoing medical
procedures, and medical practitioners), in order to provide appropriate
context-dependent behaviours and responses.

\begin{figure}[t]
    \centering
    \includegraphics[width=.6\columnwidth]{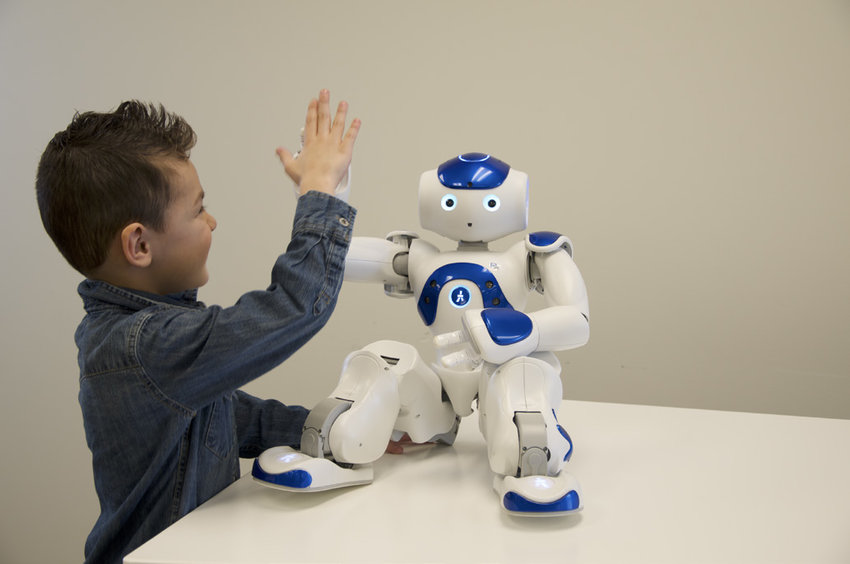}
    \caption{A child interacting with the Nao robot.}
    \label{fig:child-robot}
\end{figure}

From a technical point of view, a fundamental component in any social robot is the
action selection system which controls overall robot behaviour: the robot must make
high-level decisions as to which spoken, non-verbal, and task-based actions should be
taken next by the system as a whole. It is also crucial not only to choose the
appropriate action, but also to monitor the execution of such actions and the state
of the world detected by the robot's sensors: particularly in the context of embodied
interactions with a robot, it is likely that the predicted state will often differ
from the sensed world state, due to both the unexpected behaviour of the human
interaction partners as well as the inherent uncertainty involved in sensing and
acting in the physical world.

This paper describes a new project \cite{icsr-2020} aimed at addressing this limitation by developing
and evaluating a clinically relevant and responsive social robot, using
the Nao robot platform from SoftBank Robotics (Figure~\ref{fig:child-robot}). While
the majority of social robotics systems generally use either scripted behaviour for
action selection, or machine learning approaches to learn the correct responses to
user actions given sample inputs, we will instead use epistemic planning
techniques \cite{Dissing-Bolander:2020,Petrick-Foster:2013} as a key aspect of our approach,
as the basis for high-level action selection and execution monitoring
\cite{Petrick-Foster:2020}. In the remainder of this paper we outline the relevant
related work, proposed technical architecture, and challenges on this project.

\section{Related Work}

The area of socially assistive robotics \cite{Feil-Seifer.Mataric:2005} generally
focuses on designing a robot system with the goal of creating an effective
interaction with a human partner for the purpose of providing assistance and
achieving measurable progress in a defined domain. Such robots have been used to
improve the cognitive abilities of adult Alzheimer's patients \cite{Tapus.etal:2009},
to alleviate feelings of loneliness and depression in the elderly
\cite{Wada.etal:2004}, and to help adults with autism to improve work-related social
skills \cite{McKenna.etal:2019}. An important application of socially assistive
robots also focuses on autism in children, where robots have been used for diagnosis,
intervention, and therapy \cite{Cabibihan.etal:2013}. Since children often perceive
social robots as being similar to a companion animal or pet, such robots have also
been used for play therapy and social learning \cite{Breazeal:2011}.

Our project will explore the use of socially assistive robots for
reducing child distress and pain in clinical settings. \cite{Trost.etal:2019}
recently examined studies where a robot was used in this context:
overall, while the results seem promising and suggest that the robots succeeded in
reducing pain, a need for improved methodology and measures was identified. In
particular, the authors suggest more effective approaches could be created by
ensuring healthcare experts and system engineers collaborate from the start, and that
user and family partners contribute to a user-centred design process. Our planned
work includes input from all such groups as part of the research team
(see below). 

On the technical side, planning and interaction have a long history, and planning
techniques have been applied previously in a range of social robots and interactive systems---recent
examples include 
\cite{waldhart2016novel,sanelli2017short,kominis2017multiagent,Papaioannou2018}.
The most similar approach to ours is the JAMES social robot bartender
\cite{Petrick-Foster:2013,Petrick-Foster:2020}, which directly used an automated
planner to choose the robot's physical, sensing, and interactive actions. This system
will form the basis of the approach used on this project. Recent work on explainable
planning \cite{Fox-etal:2017} has also highlighted the links between planning and
user interaction, and is relevant to this work.

\section{Robot Behaviour and Task Environment}

The high-level goal of the project is to develop and evaluate an autonomous social
robot designed to help children deal with procedural pain in emergency room
environments. The behaviour of the robot will be based on existing cognitive
behavioural interventions that have been demonstrated to be effective in this context
(e.g., distraction, empathy). The target robot platform is the Nao robot from SoftBank
Robotics (Figure~\ref{fig:child-robot}), which has been widely used in child-robot
interaction studies, including in the identical clinical context we are
targeting \cite{Ali.etal:2019}. In addition to lab-based testing with
the robot, the system will be tested in the target environment throughout the project
period, culminating in a two-site randomised clinical trial at the end of the
project.

At the action level, the robot will engage in physical, sensory, and social actions.
Physical actions will include movement of the robot as a whole (movement between
locations, using build-in features like dancing) and moving various aspects of the
embodiment (e.g., arm and head movements).
Sensory actions will include targeted information gathering through available sensor
modalities (vision, speech) about the social context and task, but also through
verbal interaction (question asking). Social actions will include a range of dialogue
and interactive actions (question answering, engagement behaviour, explanation).
Moreover, the execution of some actions may change depending on the human
user involved and the social context (e.g., answering a question for a parent versus
a medical professional, interacting with a happy child versus one who is crying). 
The following robot utterances have been used in previous studies 
\cite{Ali.etal:2019,Trost.etal:2020} and are also relevant here:
\begin{itemize}
    \item ``I am excited to play with you today.''
    \item ``I need to catch my breath and control my breathing. Let's try this together! Breathe in 
    through your nose, hold your breath for two seconds, and out through your mouth. Try it with me!''
    \item ``I understand you are in a lot of pain right now and I’m here to help you.''
    \item ``I think we should celebrate how brave you are. Here's a new dance I just learned.''
\end{itemize}

While the robot hardware platform itself supports a range of built-in features, which
can be abstracted into high-level behaviours to support the needed actions, the
actual decision as to which behaviours are included will be decided as part of a
co-design process that involves the participation of children, parents, healthcare
professionals, as well as the technical research team. This process therefore has
implications for the technical aspects of the system design, including the process of
modelling such actions for use with the action selection and decision-making
components of the system.

\section{Overview of the Robot System}

\begin{figure}[t]
\centering
\includegraphics[width=0.90\columnwidth]{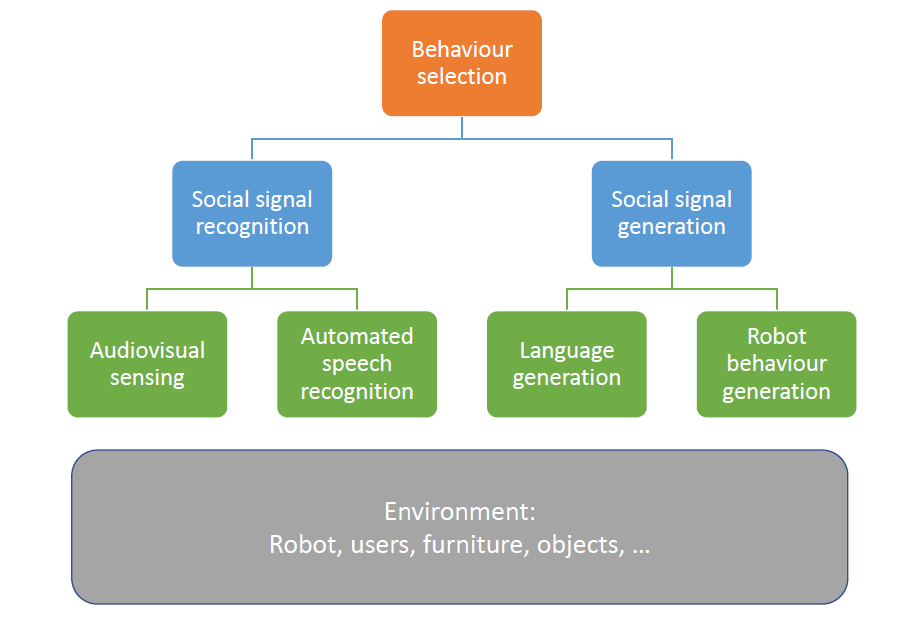}
\caption{Software architecture for the robot system.}
\label{fig:robot-architecture}
\end{figure}

The robot system will include components for social signal processing,
high-level behaviour selection, and execution monitoring and recovery as shown in
Figure~\ref{fig:robot-architecture}. The system will be based on the JAMES robot
bartender system \cite{Petrick-Foster:2013}, which uses a knowledge-level
planner to generate plans for social interaction.

\smallskip\noindent
\textbf{Social signal recognition:}
A core task in the system is to use the information from the robot's built-in
audio-visual sensors (possibly combined with environmental sensors) to determine the
state of people (children, adults) in the task area.  The particular states to be
detected will be determined through a combination of the capabilities of the sensors,
as well as the states determined to be relevant from the design process. Based on the
detected verbal and non-verbal cues employed by all humans near the robot, the state
will be estimated, including the physical state and social aspects such as attitudes,
emotions, and intention. We plan to use a neural-network approach to detect the
states, similar to the approach of \cite{Roffo.etal:2019}.

\smallskip\noindent
\textbf{Behaviour selection:}
Using the available robot actions and the detected social states from the social
signal recogniser, the system will choose appropriate high-level actions to
be performed by the robot. Action selection is performed using PKS (Planning with
Knowledge and Sensing) \cite{Petrick-Bacchus:2002,Petrick-Foster:2013}, an epistemic
planner that operates at the knowledge level and reasons about how its knowledge
state, rather than the world state, changes due to action. 
PKS has been previously used in interactive environments, such
as the JAMES robot bartender, where information-gathering dialogue actions
were modelled as sensing actions in the planner \cite{Petrick-Foster:2020}.

\smallskip\noindent
\textbf{Social signal generation:}
Once a behaviour is selected, the social signal generation process generation
converts these high-level actions into concrete robot-level actions that can be
executed on the Nao robot platform. As output, this component will produce multimodal
behaviour plans, including verbal and non-verbal actions, that are
coordinated temporally and spatially for robot execution.

\smallskip\noindent
\textbf{Execution monitoring and recovery:}
The system will monitor the changes to the state (as detected by the social signal
recogniser) while planned actions are executed on the robot, using traditional plan
monitoring techniques, but involving the richer social and epistemic states afforded
by the state recognition and planning processes. Due to the inherent uncertainty of
robot sensors and the unpredictable behaviour of humans it is likely
that the planned states will often differ from the actual world state. The
monitoring system will detect such mismatches and determine whether the
execution of the current high-level plan should continue or whether a new plan is
needed, invoking replanning as needed.

\smallskip
All software components will be developed using the Robot Operating System (ROS)
as the core middleware, together with ROSPlan
\cite{Cashmore-etal:2015} framework for integrating the planning tools.

\section{Challenges}

From a high-level planning perspective, the task of applying planning in a
child-centred medical context centres around an important knowledge engineering task
to accurately modelling the required states, actions, and goals that reflect the
types of activities the robot is expected to perform. The high-level planner is
responsible for selecting robot actions to respond appropriately in the current
social state of the task, with a mix of physical, sensory, and social behaviours.
However, unlike many robot systems which are designed from the ground up by the
technical/research team, this project involves a wider collaboration at the design
phase. Furthermore, the output of the robot may have to be
tailored to different user groups, depending on the interaction situation. Finally,
the entire robot system needs a high degree of
robustness since the user studies will include a full clinical trial. We highlight
some of these challenges below.

\smallskip\noindent
\textbf{Co-design of robot capabilities:}
At a high level, the robot must select appropriate actions to support children
undergoing clinical procedures. However, the details of the exact behaviours and
features will be developed through a co-design approach using the principles of
user-centred interaction design \cite{preece2015interaction}. The perspectives of all
groups involved in the task will be considered, including children,
parents/caregivers, healthcare professionals, as well as the research and technical
team. This process will consider all aspects of the task including the needs of
children, caregivers, and the medical team during a clinical procedure, the (possibly
differing) perceptions these groups have of the robot, and the core
functionality required for the task. While this process will be somewhat
constrained by the physical limitations of the robot platform, and the
representational and reasoning capabilities of the planner, the knowledge engineering
task of building planning models cannot be fully realised without the input from the
co-design phase.

\smallskip\noindent
\textbf{Personalised interaction:}
The robot system is meant to interact with different user groups, from children to
healthcare professionals. The kinds of plans that are generated for the various
groups will also differ, as will the potential execution of those plans on the robot.
For instance, answering a child's question may be realised differently compared with
a similar question from a parent or healthcare professional. Some of these
differences will be reflected at the action level in the planning domain model, which
also introduces some interesting connections to explainable planning (XAIP)
\cite{Fox-etal:2017}. In particular, explainability of planning decisions and plan
execution will be explored on the project, especially in the context of analysing
planner decisions by healthcare professionals.

\smallskip\noindent
\textbf{System robustness and safety:}
Several user studies are planned for the project, from initial lab-based testing
to a clinical trial, to understand the behaviour of the system. The end system is
meant to be run in a real clinical setting (at least for evaluation purposes), with
minimal human intervention. Moreover, the system is meant to run in the presence of
different user groups, including both children and adults.  As a result, the system
will require a high degree of safety and robustness at all levels.
In particular, recognising children's social signals in a real-world setting
is expected to present a particular technical challenge, 
and we will explore a range of sensor configurations and audiovisual processing approaches
to find the one that works best in this specific domain.
In  the final phase of the project, the clinical trial will evaluate the
primary goal of the project: that interaction with a robust, adaptive, socially
intelligent robot can effectively support children during a clinical procedure and
reduce their distress and pain. The clinical trial is planned for the two Canadian
paediatric departments where the co-design and usability studies will also take
place.

\section{Summary and Conclusions}

This paper describes a new research project that plans to design and use a social
robot in a medical setting to support children undergoing painful clinical procedures. At
the core of the action selection and decision-making system will be a high-level
epistemic planner that will be responsible for generating plans to interact with
different user groups, including children, parents/caregivers, and healthcare
professionals. The robot behaviour will include physical robot actions, sensing actions, and social
actions for guiding the interaction. A co-design process will be used, involving all
the user groups along with the research/technical team, which will feed into the design of the 
planning model. Some aspects of explainable planning and plan personalisation will also
be necessary on the project. The system will be demonstrated and evaluated through a
series of usability studies, leading up to a full clinical trial in a real-world
medical setting. The hope is that this work will extend existing work on social
robotics and demonstrate the potential of automated planning as a useful tool for
autonomous decision making in this challenging domain.

\section{Acknowledgements}

This work is funded by the ESRC/SSHRC Canada-UK Artificial Intelligence Initiative
through grant ES/T012986/1. We also acknowledge the lead Canadian research partners
on the project: Samina Ali, Sasha Litwin, Jennifer Parker, David Harris Smith,
Jennifer Stinson, and Frauke Zeller.

\fontsize{9.0pt}{10.0pt}\selectfont
\bibliographystyle{aaai}
\bibliography{references}

\end{document}